\documentclass[lettersize,journal]{IEEEtran}

\usepackage{times}
\usepackage{epsfig}
\usepackage{graphicx}
\usepackage{amsmath}
\usepackage{amssymb}

\usepackage{algorithm}
\usepackage{algorithmic}
\usepackage{bm}
\usepackage{amsmath}
\usepackage{amssymb}
\usepackage{threeparttable}
\usepackage{multirow}
\usepackage{graphicx}
\usepackage{subfig}
\usepackage[table]{xcolor}
\usepackage{arydshln}

\definecolor{mygray00}{gray}{.3}
\definecolor{mygray0}{gray}{.5}
\definecolor{mygray}{gray}{.85}
\definecolor{mygray1}{gray}{.9}
\definecolor{mygray2}{gray}{.95}

\newcommand{\tabincell}[2]{\begin{tabular}{@{}#1@{}}#2\end{tabular}}

\makeatletter
\newcommand{\thickhline}{%
	\noalign {\ifnum 0=`}\fi \hrule height 1pt
	\futurelet \reserved@a \@xhline
}
\makeatother

\newcommand{\app}{\raise.17ex\hbox{$\scriptstyle\sim$}}

\usepackage[pagebackref=true,breaklinks=true,letterpaper=true,colorlinks,bookmarks=false]{hyperref}

\begin{document}	
	\title{Frequency-based Matcher for Long-tailed Semantic Segmentation} 
	\author{Shan~Li, Lu~Yang, Pu~Cao, Liulei~Li, 
		and Huadong Ma$^{\star}$, \IEEEmembership{Fellow,~IEEE}
		\IEEEcompsocitemizethanks{\IEEEcompsocthanksitem 
			Shan Li and Huadong Ma (Corresponding author) are with the School of Computer Science, Beijing University of Posts and Telecommunications, Beijing, 100876, China. (e-mail: \{ls1995, mhd\}@bupt.edu.cn). \protect
			
			Lu Yang and Pu Cao are with the School of Artificial Intelligence, Beijing University of Posts and Telecommunications, Beijing, 100876, China.  (e-mail: \{soeaver,caopu\}@bupt.edu.cn).

                Liulei Li is with the University of Technology Sydney, Sydney, Australia. (e-mail: liliulei252@gmail.com).
		}
	}
	\maketitle
	
	\begin{abstract}
The successful application of semantic segmentation technology in the real world has been among the most exciting achievements in the computer vision community over the past decade.
Although the long-tailed phenomenon has been investigated in many fields, \emph{e.g.}, classification and object detection, it has not received enough attention in semantic segmentation and has become a nonnegligible obstacle to applying semantic segmentation technology in autonomous driving and virtual reality. Therefore, in this work, we focus on a relatively underexplored task setting, \textbf{long-tailed semantic segmentation} (\textbf{LTSS}). We first establish three representative datasets from different aspects, i.e., scene, object, and human. We further propose a dual-metric evaluation system and construct the LTSS benchmark to demonstrate the performance of semantic segmentation methods and long-tailed solutions. We also propose a transformer-based algorithm to improve LTSS, \textbf{frequency-based matcher}, which solves the oversuppression problem by one-to-many matching and automatically determines the number of matching queries for each class. Given the comprehensiveness of this work and the importance of the issues revealed, this work aims to promote the empirical study of semantic segmentation tasks. Our datasets, codes, and models will be publicly available\footnote{\fontsize{7pt}{1em}\url{https://github.com/caopulan/Mask2Former-LT}}.
	\end{abstract}

\section{Introduction}
\label{sec:intro}

Semantic segmentation \cite{shelhamer2016fully} allows machines to recognize images at the pixel level, which is impressive in practical applications. Owing to the continuous efforts of the community, semantic segmentation technology has advanced considerably, and new capabilities have been developed, \emph{e.g.}, domain adaptation \cite{xu2022self,ren2023prototypical}, semisupervised \cite{chen2021semi, hu2021semi}, weakly-supervised \cite{zhou2020sal, xu2023wave}, few-shot \cite{zhang2022weakly,chen2022apanet} and zero-shot semantic segmentation \cite{bucher2019zeroshot, ding2022decoupling}.

\begin{figure}
        \vspace{-.4em}
	\begin{center}
		\includegraphics[width=0.98\linewidth]{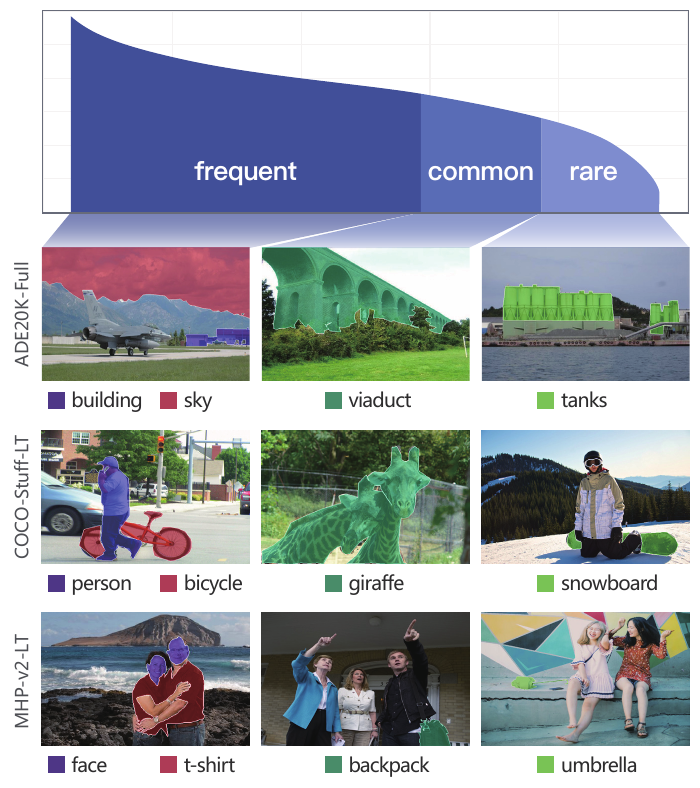}
	\end{center}
	\vspace{-0.5em}
	\caption{\textbf{Illustration of long-tailed semantic segmentation (LTSS) samples} from the three datasets we constructed. From left to right, the frequencies are frequent, common, and rare.}
	\label{fig:title}
\vspace{-5mm}
\end{figure}

As the long-tailed phenomenon \cite{chris2006longtail} prevails in the real world, it significantly constrains the applications of intelligence models. Hence, this problem has focused on several basic computer vision topics, \emph{i.e.}, classification, object detection, and instance segmentation. While semantic segmentation also represents a fundamental task, researchers have not paid enough attention to LTSS. For instance, in autonomous driving, it's essential to accurately detect and segment not only frequent objects like cars and pedestrians but also rare objects like road debris or unusual road signs, ensuring safe navigation~\cite{yang2018denseaspp,dong2020real}. 

Recently, some studies \cite{hu2021semi, wang2023balancing} have investigated this phenomenon and proposed solutions. However, most of those studies focused on insignificant long-tailed distributions in existing balanced datasets; consequently, the effectiveness of the proposed methods for LTSS cannot be fully evaluated, and targeted evaluation metrics are lacking. Based on the existing studies, we summarize three important missing points that are impeding this field: comprehensive \textbf{datasets}, a specific \textbf{evaluation system}, and an \textbf{advanced benchmark}. In this paper, we are committed to formally solving the LTSS task, building a research base to attract additional researchers.

We establish a comprehensive LTSS for empirical study. We first construct three LTSS datasets with multiple data scenarios, different long-tail degrees, and different data scales. To reveal the effects of these methods, we design a dual-metric evaluation system to measure the performance on categories with different frequencies.
Moreover, we analyze the gap between previous long-tailed solutions from other tasks and conduct experiments to demonstrate their performance. To improve the results of this task, we propose a transformer-based solution that significantly enhances the performance on low-frequency categories.

To reduce the meaningless annotation cost, we strive to reuse the existing mainstream semantic segmentation datasets to the greatest extent possible and construct three LTSS datasets: the scene-centric ADE20K-Full, the scene$\&$object-centric COCO-Stuff-LT, and the human-centric MHP-v2-LT datasets. Figure~\ref{fig:title} shows some examples of the proposed LTSS datasets. ADE20K-Full is an extended version of ADE20K \cite{zhou2017scene} and is a natural and extremely LTSS dataset. We reserved 874 classes according to \cite{cheng2021perpixel}. The other two LTSS datasets are built on balanced original collections, which hides a direct technical challenge: how can we sample in a balanced distribution to improve its long-tailedness? We propose a greedy algorithm that eliminates some data through multiple iterations. According to \cite{yang2022longtailed}, we adopt the Gini coefficient \cite{gini1912gini} to evaluate the long-tailedness of a dataset. Each iteration ensures the improvement of the Gini coefficient of the remaining data, and the iteration can be completed after the expected threshold is reached. Considering the balanced evaluation, we still use the original validation set to measure the performance. With different data scenarios, different data scales, and different long-tail degrees, the three LTSS datasets we established provide a comprehensive and unbiased evaluation of semantic segmentation models in practical settings.

Moreover, we establish a dual-metric evaluation system to evaluate the performance of LTSS models from the perspectives of easy understanding, universality, comprehensiveness, objectivity, and impartiality. Unlike image classification and object detection, semantic segmentation can measure long-tailedness at both the image and pixel levels. Thus, we divided the validation set into frequent, common, and rare splits based on the image-level and pixel-level Gini coefficients of the training set and used the mIoU metric \cite{shelhamer2016fully} for evaluation. The dual-metric evaluation system can reflect the specificity of long-tailed learning methods well and aims to provide good guidance for follow-up research.

Based on the proposed three LTSS datasets and the dual-metric evaluation system, we first evaluate two mainstream semantic segmentation models out of the box: DeepLab-V3-Plus \cite{chen2018encoder} and mask2former \cite{cheng2022masked}. To demonstrate the effects of existing long-tailed methods and illustrate the discrepancy between task settings, we evaluate three classical long-tailed learning solutions from different aspects, \emph{i.e.}, oversampling (RFS \cite{gupta2019lvis}), data augmentation (copy-paste \cite{ghiasi2021simple}) and class-level loss reweighting (seesaw loss \cite{wang2021seesaw}) under the mask2former framework, and show their effects on rare, common, and frequent categories. Experiments verify that these existing long-tailed solutions are not suitable for the LTSS task, and this task deserves additional attention in the study of dedicated methods.

Considering the challenges of LTSS, we develop a transformer-based algorithm named the \textbf{frequency-based matcher}, which introduces the multiple matching idea into bipartite graph matching \cite{carion2020end}. For semantic segmentation, one-to-one matching is unnecessary under the mask2former framework, which allows targets to match multiple queries. In our method, according to the class frequency distribution, low-frequency classes are matched to more than one query to enhance supervision. Our approach provides a new perspective, \emph{i.e.}, at the query level to balance supervision for each category with different frequencies. Extensive experiments show that our method outperforms the baseline. Due to the concise implementation of the frequency-based matcher, we recommend it as a basic component of future LTSS models.

In summary, the main contributions of our work are as follows:
\begin{enumerate}
\item This work probes the long-tailed phenomenon in semantic segmentation and is devoted to establishing the cornerstone of LTSS research from datasets, an evaluation system, and an advanced benchmark. We believe it is valuable for the community to have a formalized setting with benchmarks and evaluation metrics specifically designed for the LTSS.
\item The difference between the LTSS and previous long-tailed tasks is illustrated. We discuss the shortcomings of applying existing long-tailed solutions to LTSS and propose a targeted match-based LTSS solution.
\item Extensive experiments conducted on three LTSS datasets demonstrate the superiority of our proposed frequency-based matcher approach over classical long-tailed learning solutions. All the content will be publicly available to promote the empirical study of LTSS.
\end{enumerate}

The rest of this paper is organized as follows:
Section \ref{sec:rel} introduces related works on semantic segmentation, long-tailed learning, and transformer-based segmentation.
Section \ref{sec:ltss} introduces the dataset design, evaluation system, and advanced benchmark for the LTSS task.
The experiments and analyses are presented in Section \ref{sec:exp}. 
In Section \ref{sec:conc}, we provide the conclusion.

\section{Related Work}
\label{sec:rel}
\subsection{Semantic Segmentation}
Semantic segmentation technologies have developed rapidly since fully convolutional networks \cite{shelhamer2016fully} model dense pixel prediction. Numerous studies have focused on image context learning, \emph{e.g.}, graph models \cite{lin2016efficient}, global receptive fields \cite{chen2017deeplab}, attention mechanisms \cite{yuan2020object}, and adversarial training \cite{ yang2021attacks}, which significantly improve semantic segmentation performance. The recently proposed transformer network \cite{vaswani2017attention, dosovitskiy2020an} regards semantic segmentation as a set prediction issue and links it with instance segmentation \cite{yin2021bridging}, panoptic segmentation \cite{kirillov2019panoptic} and video segmentation \cite{xie2017object} tasks to establish a unified segmentation architecture \cite{cheng2021perpixel, cheng2022masked}, which has gradually become a new trend. In contrast, the continuous improvement of semantic segmentation technologies has led to new research directions focused on real scenarios. For example, domain adaptation semantic segmentation \cite{liu2021source} focuses on enabling a model to learn robust features across domains. Weakly-supervised semantic segmentation \cite{ahn2018learning, wang2022looking} attempts to learn dense pixel prediction with only image-level object supervision. However, few-shot semantic segmentation \cite{hu2019attention, tian2022generalized} aims to use as few samples as possible to perform segmentation. Furthermore, zero-shot semantic segmentation \cite{bucher2019zeroshot, ding2022decoupling} predicts new classes without using any labeled samples.

In contrast to these works, our study focuses on the long-tailed phenomenon \cite{chris2006longtail} in semantic segmentation, which has been largely ignored by previous researchers. The purpose is to learn dense pixel predictions that can account for both ``head" and ``tail" classes from long-tailed distributions, which will play a nonnegligible role in advancing the practical application of semantic segmentation.

\subsection{Transformer-based Segmentation}
Recently, transformers, self-attention-based neural networks originally designed for natural language processing, have achieved tremendous success in a variety of vision-processing tasks. A growing body of research shows that vision transformers can provide more powerful, uniform, and even simpler solutions for segmentation tasks.
One of the pioneering works on image segmentation is segmentation transformers (SETRs) \cite{zheng2021rethinking}. This work exploits the transformer framework to encode an image as a sequence of patches and incorporates CNN decoders to increase feature resolution.
As a follow-up to SETR, segmenter \cite{strudel2021segmenter} is an encoder-decoder architecture built on a pure transformer approach for semantic segmentation. Different from the SETR, the segmenter decodes with a mask transformer and incorporates label information into the decoder for modeling.
To overcome the efficiency limitations of visual transformers, SegFormer \cite{xie2021segformer} designed a lightweight and hierarchically structured transformer encoder with a multilayer perceptron (MLP) decoder, which jointly ensures speed, accuracy, and robustness for semantic segmentation. Some work has been done to improve the decoder \cite{cheng2021perpixel, zhang2021k}. For example, mask2former \cite{cheng2022masked} replaces the cross-attention used in the standard transformer decoder with masked cross-attention, which allows the query to focus on only localized features centered around predicted segments. Subsequent studies \cite{chen2022group, li2023mask, zhang2023mpformer} have improved upon the abovementioned components of mask2former, making transformer-based segmentation the mainstream and best-performing segmentation framework at present.

\subsection{Long-tailed Learning}
In nature or real life, a distribution of random variables is more widespread than the uniform distribution, \emph{i.e.}, the long-tailed distribution \cite{chris2006longtail}. As a result, models naively trained on long-tailed data perform significantly worse on tail classes than on head classes. This phenomenon first attracted attention in image classification tasks \cite{van2018inaturalist, liu2019large}, and a variety of methods have been developed to improve the recognition performance of tail classes. Oversampling \cite{shen2016relay, kim2020m2m} and data augmentation \cite{chou2020remix, ghiasi2021simple} techniques are adopted to increase the number of samples for tail data to achieve a balanced learning effect. Cost-sensitive reweighting \cite{cui2019class, wang2021seesaw} aims to modify the gradient to improve the modeling ability of tail classes by assigning weights to different classes or hard examples. The LVIS \cite{gupta2019lvis} dataset introduces a new direction for the community: long-tailed object detection/instance segmentation leading to a series of new problems. For example, tail classes are easily recognizable as background signals, resulting in missed detection \cite{dave2021eval}. Alternatively, sparse tail samples are difficult to distinguish from background signals through naive learning, resulting in inaccurate masking of tail classes \cite{zhang2021refinemask}.

Although long-tail learning has attracted increasing attention, it has obvious shortcomings in semantic segmentation, which is another basic task of visual recognition in addition to image classification and object detection. Moreover, previous studies have shown that differences in learning architecture and supervision form caused by different task settings can also lead to new long-tailed problems. Therefore, communities urgently need to establish long-tailed learning systems for semantic segmentation tasks and propose targeted solutions.

\section{Long-tailed Semantic Segmentation}
\label{sec:ltss}
In this section, we first formulate our problem mathematically in Section \ref{sec:ltss-pf}. Then, we specify the LTSS dataset construction process and the dataset statistics in Section \ref{sec:ltss-dd}. In Section \ref{sec:ltss-evalsys}, we introduce the evaluation system. Finally, in Section \ref{sec:ltss-methods}, we propose the frequency-based matcher method for the LTSS task and compare it with classical long-tailed solutions.

\subsection{Problem Formulation}
\label{sec:ltss-pf}
Formally, semantic segmentation aims to establish a mapping $\textit{f}$: $\bm{\mathcal{X}}\mapsto\bm{\mathcal{Y}}$, where \bm{$\mathcal{X}$} is the input image space and \bm{$\mathcal{Y}$} is the output dense pixel segmentation space. $\mathcal{C}$ classes are predefined to describe each sample in \bm{$\mathcal{Y}$}. When the \bm{$\mathcal{Y}$} space follows the long-tailed distribution on $\mathcal{C}$ classes, solving the mapping $\textit{f}$ is an LTSS problem.

\subsection{Dataset Design}
\label{sec:ltss-dd}
During the construction of the LTSS dataset, we first propose measuring the dataset's long-tailedness at both the image and pixel levels. Then, a greedy algorithm is introduced to construct LTSS datasets from off-the-shelf balanced datasets. As a result, three LTSS datasets are built, and the statistics of each LTSS dataset are described in detail.

\noindent\textbf{Measure the Long-tailedness.} The precondition for building a long-tailed dataset is how to measure the dataset's long-tailedness. According to \cite{yang2022longtailed}, the Gini coefficient is an effective and objective evaluation metric. However, unlike image classification and object detection datasets, semantic segmentation datasets assign labels to pixels rather than images/instances. Therefore, calculating the weights of different classes is not intuitive. Therefore, we propose two methods for calculating class weights at the image and pixel levels, which are used to measure the long-tailedness of semantic segmentation datasets and guide subsequent research. Specifically, for class $c \!\in\! \{1, ..., \mathcal{C}\}$, the image-level weight in the dataset is:
\begin{equation}
\begin{aligned}
\mathcal{P}^\textit{image}_{\textit{c}} = \sum\limits_{i=1}^{N} \varepsilon(\mathcal{S}^\textit{c}_\textit{i}),
\end{aligned}
\label{eq:image}
\end{equation}
where $\mathcal{S}^\textit{c}_\textit{i}$ is the pixel number of class $c$ in image $\textit{i}$ and $\varepsilon$ is a binary function: $\varepsilon(x) = 0$ (when $x = 0$) or $\varepsilon(x) = 1$ (when $x > 0$),
\emph{i.e.}, $\mathcal{P}^\textit{image}_{\textit{c}}$ represents the frequency of class $c$ in the dataset. Similarly, the pixel-level weight for $c \!\in\! \{ 1, ..., \mathcal{C} \}$ is:
 \begin{equation}
\begin{aligned}
\mathcal{P}^\textit{pixel}_{\textit{c}} = \sum\limits_{i=1}^{N} \frac{\mathcal{S}^\textit{c}_\textit{i}}{H_\textit{i} * W_\textit{i}},
\end{aligned}
\label{eq:pixel}
\end{equation}
$H_\textit{i}$ / $W_\textit{i}$ is the height/width of image $\textit{i}$, \emph{i.e.}, $\mathcal{P}^\textit{pixel}_{\textit{c}}$ represents the sum of the pixel proportions of class $c$ in each image. Although both $\mathcal{P}^\textit{image}_{\textit{c}}$ and $\mathcal{P}^\textit{pixel}_{\textit{c}}$ can measure the class weight in the dataset, they reflect different characteristics of objects (or stuffs). For example, objects with low diversity tend to have smaller $\mathcal{P}^\textit{image}_{\textit{c}}$, and objects with smaller scales tend to have smaller $\mathcal{P}^\textit{pixel}_{\textit{c}}$. Based on $\mathcal{P}^\textit{image}_{\textit{c}}$ and $\mathcal{P}^\textit{pixel}_{\textit{c}}$, we can obtain the distributions $\mathcal{D}^\textit{image}$ and $\mathcal{D}^\textit{pixel}$, where $\mathcal{D^\textit{image / pixel}}=\{\mathcal{P}^\textit{image / pixel}_{c}\}_{c=1}^{\mathcal{C}}$, and calculate the image-level Gini coefficient $\delta^\textit{image}$ and pixel-level Gini coefficient $\delta^\textit{pixel}$ of each semantic segmentation dataset.

\begin{table*}[htbp]
	\centering
	\begin{threeparttable}
	\resizebox{1\textwidth}{!}{
	\setlength\tabcolsep{5pt}
	\renewcommand\arraystretch{1.3}
	\begin{tabular}{c|c|c|c|c|c|c|c|c|c|c|c}
	\hline\thickhline
	\rowcolor{mygray1}
	                                       &                                           &                                                      &                                        &                                          &                                                 &         \multicolumn{3}{c|}{Image-level}       &         \multicolumn{3}{c}{Pixel-level}         \\
	\cline{7-9} \cline{10-12}
	\rowcolor{mygray1}
        \multirow{-2}{*}{Dataset} & \multirow{-2}{*}{\#Images} & \multirow{-2}{*}{\#Train/Val/Test/} & \multirow{-2}{*}{\#Class} & \multirow{-2}{*}{Purpose} & \multirow{-2}{*}{Anno. Types}  & Max Size & Min Size & $\delta^\textit{image}$ & Max Size & Min Size & $\delta^\textit{pixel}$ \\
	\hline
	\hline
	\textcolor{mygray0}{ImageNet-LT \cite{ammirato2017dataset}} & \textcolor{mygray0}{185,846} & \textcolor{mygray0}{115,846/20,000/50,000}  & \textcolor{mygray0}{1,000}& \textcolor{mygray0}{Object} & \textcolor{mygray0}{Cls}  & \textcolor{mygray0}{1,280} & \textcolor{mygray0}{5} & \textcolor{mygray0}{0.524} & \textcolor{mygray0}{-} & \textcolor{mygray0}{-} & \textcolor{mygray0}{-} \\
	\textcolor{mygray0}{Places-LT \cite{liu2019large}} & \textcolor{mygray0}{106,300} & \textcolor{mygray0}{62,500/7,300/36,500}  & \textcolor{mygray0}{365}& \textcolor{mygray0}{Scene} & \textcolor{mygray0}{Cls}  & \textcolor{mygray0}{4,980} & \textcolor{mygray0}{5} & \textcolor{mygray0}{0.671} & \textcolor{mygray0}{-} & \textcolor{mygray0}{-} & \textcolor{mygray0}{-} \\
	\textcolor{mygray0}{LVIS-v1 \cite{gupta2019lvis}} & \textcolor{mygray0}{159,623} & \textcolor{mygray0}{100,170/19,809/39,644}  & \textcolor{mygray0}{1,203}& \textcolor{mygray0}{Object} & \textcolor{mygray0}{BBox/Mask}  & \textcolor{mygray0}{50,552} & \textcolor{mygray0}{1} & \textcolor{mygray0}{0.820} & \textcolor{mygray0}{-} & \textcolor{mygray0}{-} & \textcolor{mygray0}{-} \\
	\hline
	\hline
	\textcolor{mygray0}{ADE20K \cite{zhou2017scene}} & \textcolor{mygray0}{25,562} & \textcolor{mygray0}{20,210/2,000/3,352}  & \textcolor{mygray0}{150}& \textcolor{mygray0}{Scene} & \textcolor{mygray0}{SemSeg}  & \textcolor{mygray0}{11,588} & \textcolor{mygray0}{41} & \textcolor{mygray0}{0.645} & \textcolor{mygray0}{3,014.9} & \textcolor{mygray0}{3.84} & \textcolor{mygray0}{0.801}  \\
	ADE20K-Full \cite{cheng2021perpixel}& 27,574 & 25,574/2,000/-  & 847 & Scene & SemSeg & 13,445 & 1 & 0.865 & 3,466.1 & 0.0006 & 0.934 \\
        \hline
	\textcolor{mygray0}{COCO-Stuff \cite{caesar2018cocostuff}} & \textcolor{mygray0}{163,957} & \textcolor{mygray0}{118,287/5,000/40,670}  & \textcolor{mygray0}{171}& \textcolor{mygray0}{Object} & \textcolor{mygray0}{SemSeg}  & \textcolor{mygray0}{63,965} & \textcolor{mygray0}{121} & \textcolor{mygray0}{0.517} & \textcolor{mygray0}{10,021.1} & \textcolor{mygray0}{3.82} & \textcolor{mygray0}{0.653}  \\
	COCO-Stuff-LT (ours) & 87,614 & 40,679/5,000/40,670  & 171 & Object & SemSeg & 23,557 & 2 & 0.669 & 3,491.0 & 0.014 &  0.773 \\
	\hline
	\textcolor{mygray0}{MHP-v2 \cite{zhao2018understanding}} & \textcolor{mygray0}{25,403} & \textcolor{mygray0}{15,403/5,000/5,000}  & \textcolor{mygray0}{59}& \textcolor{mygray0}{Human} &  \textcolor{mygray0}{SemSeg}  & \textcolor{mygray0}{15,403} & \textcolor{mygray0}{44} & \textcolor{mygray0}{0.601} & \textcolor{mygray0}{9,733.8} & \textcolor{mygray0}{0.823} & \textcolor{mygray0}{0.885}  \\
	MHP-v2-LT (ours) & 16,931 & 6,931/5,000/5,000  & 59 & Human & SemSeg & 6,931 & 1 & 0.701 & 4,279.4 & 0.004 &  0.909 \\
	\hline
	\end{tabular}
	}
	\end{threeparttable}
	\caption{\textbf{Statistics of LTSS datasets}. Black denotes the three LTSS datasets established in our work.  Here, ``Anno. Types'' means annotation types. $\delta^\textit{image}$ and $\delta^\textit{pixel}$ are the image-level Gini coefficient and  the pixel-level Gini coefficient, respectively.}
	\label{table:ltss-datasets}
\end{table*}

\noindent\textbf{Sample in a Balanced Distribution.} As shown in Table~\ref{table:ltss-datasets}, most of the existing semantic segmentation datasets (\emph{e.g.}, ADE20K \cite{zhou2017scene}, COCO-Stuff \cite{caesar2018cocostuff} and MHP-v2 \cite{zhao2018understanding}) exhibit a degree of long-tailedness at the image level. However, these datasets still contain fewer images than long-tailed datasets for image classification and object detection (\emph{e.g.}, ImageNet-LT \cite{ammirato2017dataset}, Places-LT \cite{liu2019large}, and LVIS-v1 \cite{gupta2019lvis}). Gratifyingly, we find that ADE20K-Full \cite{zhou2017scene, cheng2021perpixel}, an extended version of ADE20K, is a natural long-tailed semantic segmentation dataset with significant long-tailedness both at the image level and at the pixel level. Therefore, we consider ADE20K-Full as an LTSS dataset and recommend it as the primary benchmark.

Using only one dataset is insufficient for comprehensively measuring the effectiveness of long-tailed algorithms; therefore, we hope to construct several diverse low-cost LTSS datasets. Inspired by \cite{liu2019large}, sampling a long-tailed subset with a distribution $\mathcal{D}^\textit{image/pixel}_{\textit{sub}}$ in an off-shelf balanced dataset is an effective method. However, this is a complex optimization problem because each sample in the LTSS dataset has multiple labels, and it is difficult to find an intuitive method for obtaining a subset of controllable distributions. ImageNet-LT and Places-LT are single-label datasets, and LVIS-v1 adopts the federated dataset design, cleverly avoiding this problem. To this end, we use a concise but effective greedy algorithm that continuously improves the Gini coefficient of the remaining subsets by eliminating some data iteratively.

The image-level Gini coefficient is taken as an example.
First, we deduce a distribution $\mathcal{D}^\textit{image}_{\textit{tgt}}$ according to the expected Gini coefficient $\delta^\textit{image}_{\textit{tgt}}$.
Second, starting from the class $c$ with the smallest image-level weight $\mathcal{P}^\textit{image}$ in the original distribution, we select all the samples containing class $c$ and remove part of the samples, making the image-level weight of class $c$ of the remaining samples close to the target distribution $\mathcal{D}^\textit{image}_{\textit{tgt}}$. We mark the remaining samples as reserved data (to avoid eliminating these samples in subsequent iterations, resulting in an image-level weight of class $c$ being 0) and obtain a new distribution $\mathcal{D}^\textit{image}_{\textit{i}}$. Then, if the Gini coefficient $\delta^\textit{image}_{\textit{i}}$ of distribution $\mathcal{D}^\textit{image}_{\textit{i}}$ is greater than or equal to $\delta^\textit{image}_{\textit{tgt}}$, the iteration stops. Otherwise, we select the class $c$ with the second smallest weight in the original distribution and repeat the above process until all classes are iterated. We also set a threshold $\mathcal{T}$ to prevent too many samples from being eliminated. The iteration also stops when the number of eliminated samples is greater than $\mathcal{T}$.

Based on the above process, we sample one subset from the COCO-Stuff and MHP-v2 datasets, denoted as COCO-Stuff-LT and MHP-v2-LT, respectively. Together these two datasets with ADE20K-Full, are the three LTSS datasets built by our work.

\begin{figure*}[htbp]
        \vspace{-.6em}
	\centering
	\subfloat[ADE20K-Full at image-level]{
		\begin{minipage}[t]{0.32\linewidth}
			\centering
			\includegraphics[height=1.7in]{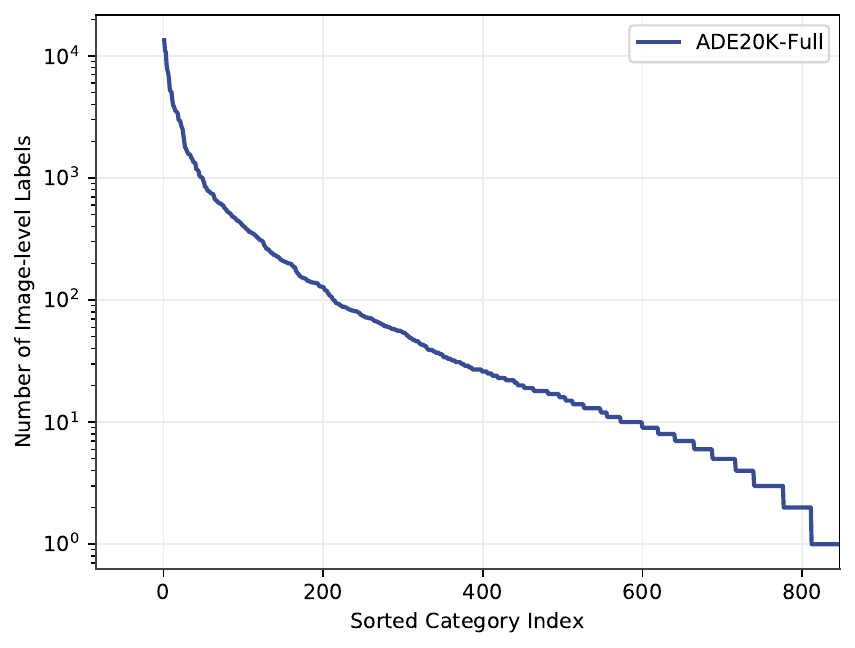}
	\end{minipage}}
	\subfloat[COCO-Stuff-LT at the image level]{
		\begin{minipage}[t]{0.32\linewidth}
			\centering
			\includegraphics[height=1.7in]{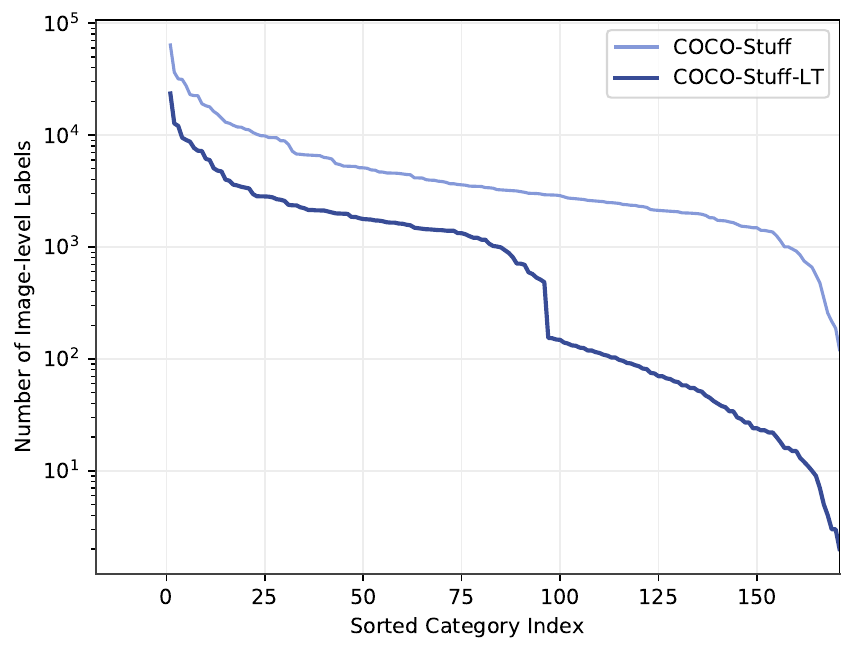}
	\end{minipage}}
	\subfloat[MHP-v2-LT at image-level]{
		\begin{minipage}[t]{0.32\linewidth}
			\centering
			\includegraphics[height=1.7in]{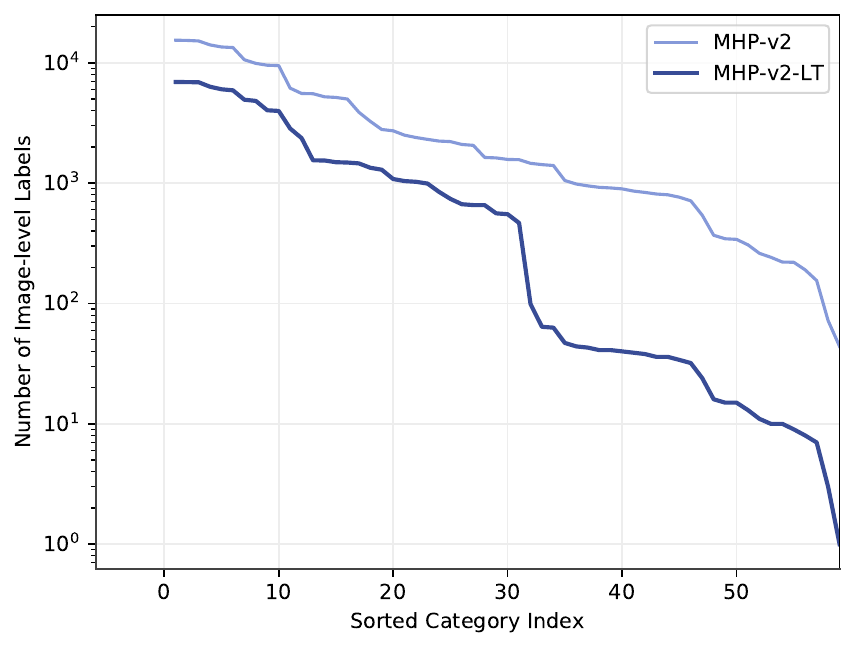}
	\end{minipage}}
	\\
	\subfloat[ADE20K-Full-LT at the pixellevel]{
		\begin{minipage}[t]{0.32\linewidth}
			\centering
			\includegraphics[height=1.7in]{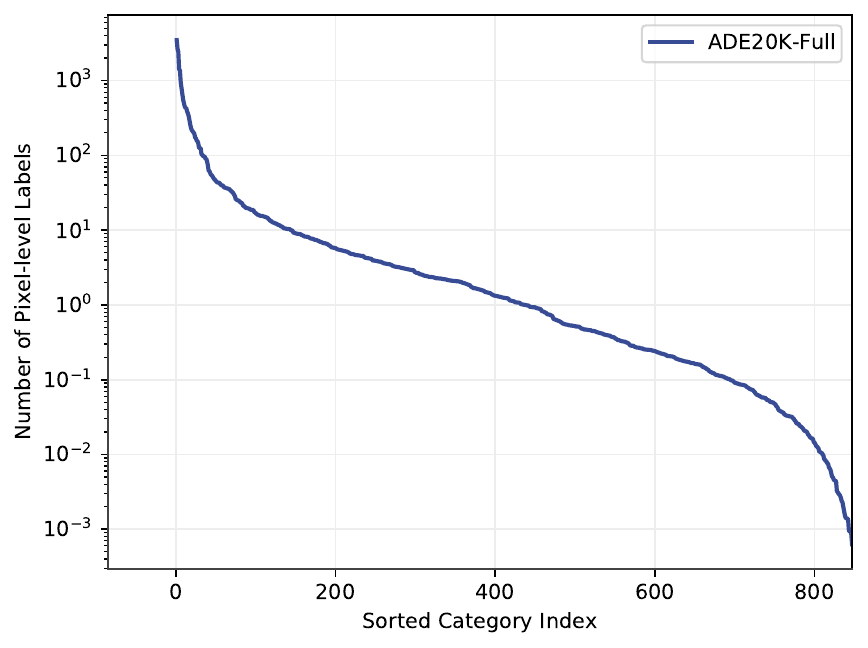}
	\end{minipage}}
	\subfloat[COCO-Stuff-LT at the pixel level]{
		\begin{minipage}[t]{0.32\linewidth}
			\centering
			\includegraphics[height=1.7in]{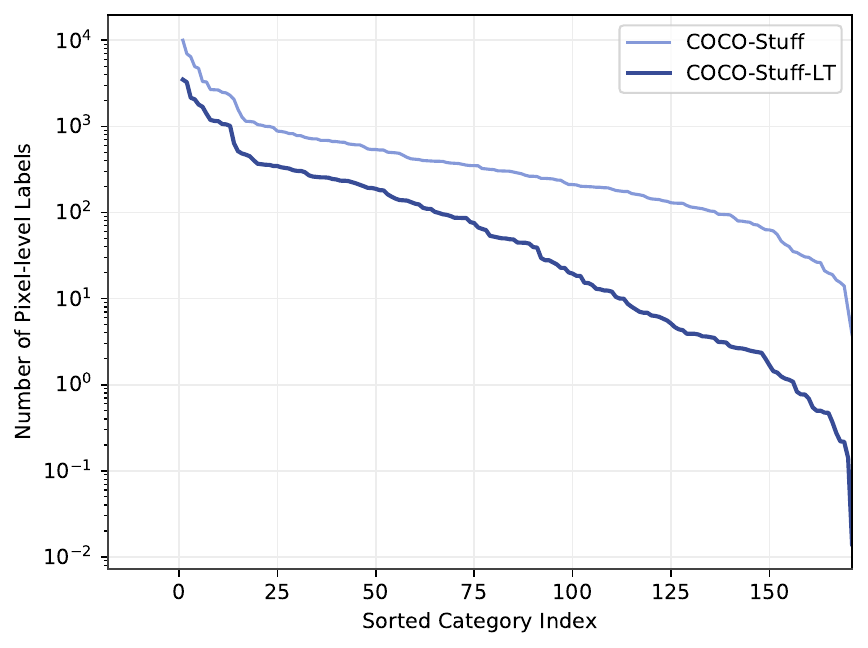}
	\end{minipage}}
	\subfloat[MHP-v2-LT at the pixel level]{
		\begin{minipage}[t]{0.32\linewidth}
			\centering
			\includegraphics[height=1.7in]{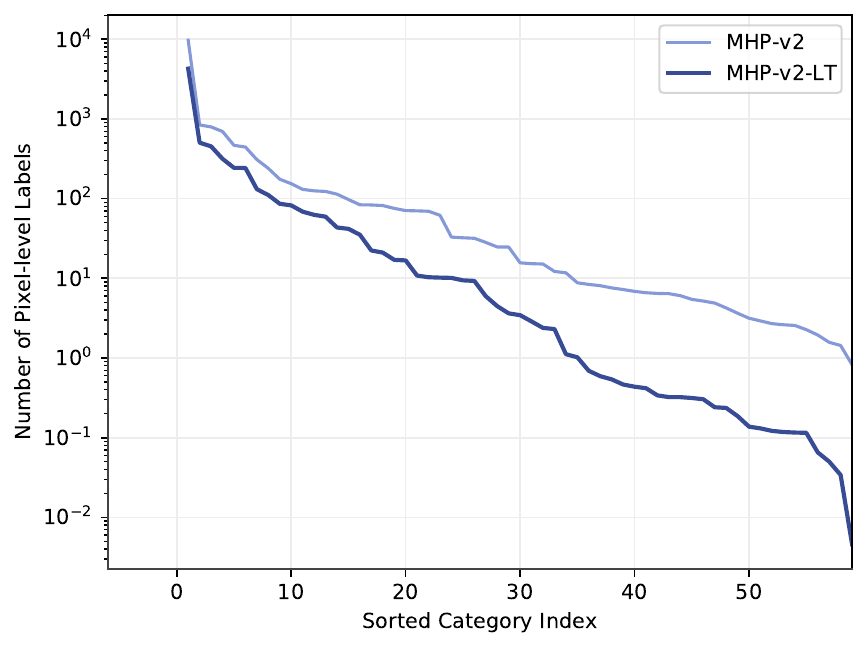}
	\end{minipage}}
	\caption{\textbf{The label distribution of different LTSS datasets}. Figures (a)-(c) in the 1$st$ row show the long-tailed label distributions at the image level, and Figures (d)-(f) in the 2$nd$ row show the long-tailed label distributions at the pixel level.}
	\label{long-tailed_curves}
\vspace{-5mm}
\end{figure*}

\noindent\textbf{Dataset Statistics.} Table~\ref{table:ltss-datasets} shows several important statistics of the three constructed LTSS datasets. Figure \ref{long-tailed_curves} illustrates the label distribution of these LTSS datasets at both the image and pixel levels. In terms of the number of images, these three datasets have obvious differences, especially in terms of the number of samples in the training set. COCO-Stuff-LT has the largest training set, approximately 40,679 images, equivalent to \app35\% of the original COCO-Stuff, while ADE20K-Full and MHP-v2-LT are followed by 25,574 and 6,931 training set samples, respectively. In terms of the class number, ADE20K-Full ranks first with 847 classes, including frequent classes such as \texttt{building} and \texttt{sky}, and rare classes such as \texttt{sword} and \texttt{snowboard}. This is also the reason why ADE20K-Full is a natural long-tailed dataset. COCO-Stuff-LT and MHP-v2-LT are the LTSS datasets generated in this work. We retain the class number of their original versions (171 and 59 classes), forming a significant difference from ADE20K-Full, which will be conducive to measuring the performance of the LTSS algorithm at different class scales. A more significant feature is that these three LTSS datasets are based on different scenarios: ADE20K-Full is scene-centric, COCO-Stuff-LT is both scene- and object-centric, and MHP-v2-LT is human-centric, basically covering the current mainstream wild scenarios.

Regarding long-tailedness, the $\delta^\textit{image}$ of ADE20K-Full is 0.865, approximately \app34\% higher than that of the original ADE20K (0.645). The $\delta^\textit{pixel}$ of ADE20K-Full reached an amazing value of 0.934, indicating many small-scale objects or objects in it. The Gini coefficients of COCO-Stuff-LT and MHP-v2-LT are also significantly improved compared to those of the original versions, proving that our proposed greedy algorithm can effectively sample a long-tailed subset in the balanced dataset. We can more intuitively observe the difference in the label distributions between the LTSS and balanced datasets. This difference is mainly reflected in the tail classes, which have a similar number of images.

\subsection{Evaluation System}
\label{sec:ltss-evalsys}
Long-tailed image classification and object detection have similar evaluation principles. For example, long-tailed image classification divides all the classes into three groups according to the label distribution of the training set: many-shot (over 100 images), medium-shot (21\app100 images), and few-shot (1\app20 images); additionally, the top-1 classification accuracy is calculated for each group.
For long-tailed object detection/instance segmentation, taking the LVIS-v1 dataset as an example, 1,203 classes are divided into three buckets based on the number of images in which the class is annotated: frequent (over 100 images), common (11\app100 images), and rare (1\app10 images).
The average precision (AP) \cite{lin2014microsoft} is calculated for all three buckets.

Considering the difference in the label distributions of the three LTSS datasets, we change the fixed threshold to a relative proportion, \emph{i.e.,}, the first 60\% of $\mathcal{D}$ is frequent, the middle 20\% is common, and the last 20\% is rare. Based on $\mathcal{D}^\textit{image}$ and $\mathcal{D}^\textit{pixel}$, two division modes, image-level and pixel-level, are formed. The classic mean intersection over union (mIoU) \cite{shelhamer2016fully} is used as the evaluation metric and is calculated independently of different splits.
This is the proposed dual-metric evaluation system for the LTSS algorithm, and it is employed in our experiments.

\subsection{Advanced Benchmark}
\label{sec:ltss-methods}

In this section, we first revisit classical long-tailed solutions and analyze the gap between them and the LTSS task. Three are evaluated based on the proposed LTSS datasets and evaluation system. We then propose a new transformer-based method, frequency-based matcher, to address the LTSS problem. These classical long-tailed solutions and our proposed frequency-based matcher method are recommended for use as benchmarks for follow-up LTSS researchers.

\noindent\textbf{Revisiting Classical Solutions.} At present, long-tailed learning has formed several mainstream solutions, \emph{e.g.}, data processing methods, cost-sensitive weighting, decoupled representation, and transfer learning. These classical solutions attempt to solve the imbalance between head and tail classes from different perspectives of representation learning and have succeeded in long-tailed image classification and object detection. To observe the adaptability of classical long-tailed learning methods to semantic segmentation frameworks \cite{cheng2022masked} and the similarities and differences between LTSS and other long-tailed tasks, we select the following three representative classical long-tailed learning methods from different aspects (\emph{i.e.}, sampling strategy, data augmentation, and loss function) to verify their impact on LTSS.

We then investigate the sampling strategy, data augmentation, and loss function. (1) Repeat factor sampling (RFS) \cite{gupta2019lvis} is an oversampling method in which a rebalancing operation is performed by increasing the sampling frequency of images containing tail classes; this approach is a widely used and effective solution in long-tailed object detection/instance segmentation \cite{hsieh2021droploss, li2022eod}. (2) Copy-Paste \cite{dvornik2018modeling, ghiasi2021simple} is a mixed image augmentation method that expands training samples by copying instances from one image to another. (3) The seesaw loss \cite{wang2021seesaw} adopts a gradient-guided reweighing mechanism. The ratio of accumulated positive gradients to negative gradients is used to independently increase the weight of positive gradients and reduce the weight of negative gradients for each classifier. 
The above three solutions were combined with mask2former. Hyperparameter tuning is adopted for fair results.

Notably, as efficiently learning from long-tailed distributions has been fully studied in image classification \cite{liu2019large, park2022cmo} and object detection/instance segmentation tasks \cite{gupta2019lvis, tan2020equalization}, these studies have gaps in the LTSS task. First, semantic segmentation is a pixel-level visual recognition task, and existing long-tailed solutions at the image and region levels are not applicable. Second, there is unique prior knowledge (\emph{e.g.}, scene graphs \cite{yoon2021image, li2022the}, object relations \cite{zhou2020ccascaded, li2022improving} and human body structures \cite{yang2019parsing, yang2022part, yang2023deep}) that can help tail class recognition in semantic segmentation application scenarios; however, this requires the development of targeted solutions. Hence, LTSS-specific solutions are still needed.


\noindent\textbf{Frequency-based Matcher.}
Although the aforementioned classical solutions are designed to solve the long-tailed problem (\emph{i.e.}, solving long-tailed instance segmentation on LVIS \cite{gupta2019lvis}), discrepancies between datasets and tasks may weaken their effects. Specifically, LVIS is a federated dataset, indicating that resampling and copy-paste strategies can directly increase low-frequency class instances. However, when applied to proposed LTSS datasets, these methods also cause high-frequency classes to repeat. Hence, a crucial problem is enhancing supervision from only low-frequency categories in the LTSS task.

\begin{figure}
	\begin{center}
		\includegraphics[width=0.99\linewidth]{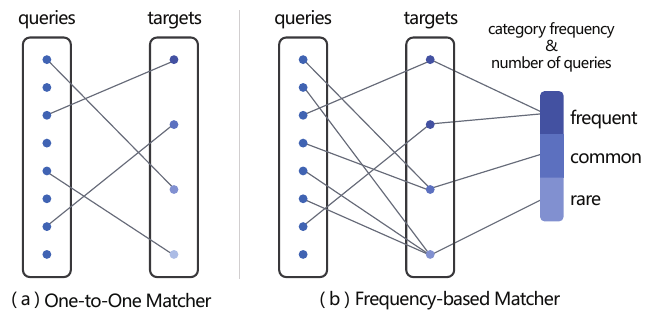}
	\end{center}
\vspace{-5mm}
	\caption{\textbf{Illustration of the proposed frequency-based matcher}. (a) shows the one-to-one matcher in \cite{cheng2022masked}. (b) The proposed frequency-based matcher in this paper.}
	\label{frequency-based_matcher}
	\vspace{-5mm}
\end{figure}

Taking mask2former as the baseline, we find that improving supervision from low-frequency classes at the query level is possible. Let us revisit the matching and inference process in mask2former first. During training, as the cost matrix between all queries and targets is acquired, the Hungarian algorithm finds an optimal assignment that follows a one-to-one paradigm (Figure~\ref{frequency-based_matcher} (a)). If there are $m$ queries and $n$ targets ($n\leq m$), $n$ queries will be matched to $n$ targets, while the other $m-n$ queries will receive no-object. This process produces multiple negative samples since $m$ is usually much larger than $n$. The supervision from low-frequency categories is diminished to a great extent due to the large number of no-object targets and other higher-frequency category targets. Moreover, the inference process of semantic segmentation follows $M_c=\sum_{i=1}^{N}p_i^c\times m_i$, where $M_c$, $p_i^c$, and $m_i$ are the $c$th class mask, probability of the $c$th class from the $i$th query, and $i$th query's mask, respectively. Different from instance-level tasks (\emph{i.e.}, detection and instance segmentation), prediction results are obtained from all queries instead of from responsive queries, indicating that one-to-one matching is not indispensable in the semantic segmentation task.

Hence, we use a frequency-based matcher to enhance supervision from low-frequency classes. Specifically, we design a one-to-many matching strategy assigning more queries for low-frequency classes during training. Like \cite{gupta2019lvis}, for each class $c$ with frequency $p_c=\mathcal{P}_c/N$, we define the class-level query number as follows:
\begin{equation}
\begin{aligned}
q_c=\left \lceil s\times max(1,\sqrt{\frac{t}{p_c}})\right\rceil,
\end{aligned}
\label{eq:freq}
\end{equation}
where $t$ and $s$ are two hyperparameters. In particular, $t$ is a frequency-related threshold that controls the relative matching intensity between all classes. A class with a frequency less than $t$ matches more than one query, and the intensity increases as the frequency decreases, as shown in Figure~\ref{frequency-based_matcher} (b). The additional hyperparameter $s$ controls the overall matching intensity during training, as we mentioned that one-to-one matching is not indispensable in LTSS.

\noindent\textbf{Compared to Classical Solutions.} Our frequency-based matcher is more effective and suitable for LTSS. First, our query-level mechanism can only \emph{repeat} or \emph{copy} low-frequency targets. We solve the abovementioned problems associated with resampling at the image level (RFS) or region level (copy-paste). Moreover, compared to the long-tailed loss reweighting method (seesaw loss), our one-to-many matching approach provides greater diversity of gradient information. Loss reweighting directly amplifies the low-frequency classification loss, while our method matches multiple queries with a low-frequency target and calculates the loss separately. Our frequency-based matcher is specifically designed for LTSS, and extensive experiments will demonstrate its superiority.

\begin{table*}[h!]
\centering
\setlength\tabcolsep{2.0pt}
\renewcommand\arraystretch{1.3}
\resizebox{1.0\textwidth}{!}
{
\begin{tabular}{l|c|c|c|c|c|c|c|c|c|c|c|c|c}
\hline\thickhline
\rowcolor{mygray1}
 & \multicolumn{3}{c|}{ADE20K-Full}       &         \multicolumn{3}{c|}{COCO-Stuff-LT} &         \multicolumn{3}{c|}{MHP-v2-LT} &  & \multicolumn{3}{c}{Classical Datasets} \\ 
\cline{2-10}  \cline{12-14}  
\rowcolor{mygray1}
\multirow{-2}{*}{Method} & mIoU & Image. & Pixel. & mIoU & Image. & Pixel. & mIoU & Image. & Pixel. & \multirow{-2}{*}{$\Delta_{freq}$} & ADE20K & COCO-Stuff & MHP-v2\\     \hline \hline
\multirow{3}{*}{DeepLab-V3-Plus \cite{chen2018encoder}} & \multicolumn{1}{c|}{\multirow{3}{*}{13.6}} & 1.5 & 0.2 & \multirow{3}{*}{27.7} & 8.0 & 7.5 & \multirow{3}{*}{26.9} & 4.0 & 5.3 & \multirow{3}{*}{-} & \multirow{3}{*}{41.2} & \multirow{3}{*}{40.8} & \multirow{3}{*}{39.8} \\
& \multicolumn{1}{c|}{} & 7.6 & 2.1 & & 18.4 & 15.7 & & 5.1 & 4.8 & & & \\
& \multicolumn{1}{c|}{} & 19.4 & 21.8 & & 37.2 & 38.3 & & 41.1 & 40.8 & & & \\ \hline\hline
\multirow{3}{*}{Mask2Former \cite{cheng2022masked}} & \multirow{3}{*}{18.8} & 4.8 & 3.5 & \multirow{3}{*}{32.9} & 14.2 & 13.8 & \multirow{3}{*}{32.3} & 8.8 & 13.8 & \multirow{3}{*}{-} & \multirow{3}{*}{47.2} & \multirow{3}{*}{46.5} & \multirow{3}{*}{44.6} \\
& \multicolumn{1}{c|}{} & 13.4 & 6.2 & & 24.8 & 21.0 & & 10.4 & 10.6 &  & & \\
& \multicolumn{1}{c|}{} & 25.1 & 28.1 & & 41.7 & 43.0 & & 46.8 & 45.4 & & & \\ \hdashline[3pt/2pt]
\multirow{3}{*}{+ RFS \cite{gupta2019lvis}} & \multirow{3}{*}{19.1} & 5.7 \textcolor{red}{(+0.9)} & 4.1 \textcolor{red}{(+0.6)} & \multirow{3}{*}{34.2} & 16.1 \textcolor{red}{(+1.9)} & 15.0 \textcolor{red}{(+1.2)} & \multirow{3}{*}{32.6} & 10.0 \textcolor{red}{(+1.2)} & 14.5 \textcolor{red}{(+0.7)} & \textcolor{red}{+1.08} & \multirow{3}{*}{-} & \multirow{3}{*}{-} & \multirow{3}{*}{-} \\
& & 13.9 \textcolor{red}{(+0.5)} & 6.5 \textcolor{red}{(+0.3)} & & 27.0 \textcolor{red}{(+2.2)} & 24.0 \textcolor{red}{(+3.0)} & & 9.5 \textcolor[rgb]{0.17,0.62,0.17}{(-0.9)} & 10.1 \textcolor[rgb]{0.17,0.62,0.17}{(-0.5)}& \textcolor{red}{+0.76} & & & \\
&  & 25.2 \textcolor{red}{(+0.1)} & 28.3 \textcolor{red}{(+0.2)} & & 42.6 \textcolor{red}{(+0.9)} & 43.9 \textcolor{red}{(+0.9)} & & 47.1 \textcolor{red}{(+0.3)} & 45.5 \textcolor{red}{(+0.1)} & \textcolor{red}{+0.41} & & & \\ \hdashline[3pt/2pt]
\multirow{3}{*}{+ Copy-Paste \cite{ghiasi2021simple}} & \multirow{3}{*}{19.7} & 4.8 (+0.0) & 3.7 \textcolor{red}{(+0.2)} & \multirow{3}{*}{34.3} & 14.6 \textcolor{red}{(+0.4)} & 13.5 \textcolor[rgb]{0.17,0.62,0.17}{(-0.3)} & \multirow{3}{*}{33.1} & 9.2 \textcolor{red}{(+0.4)} & 13.4 \textcolor[rgb]{0.17,0.62,0.17}{(-0.4)} &  \textcolor{red}{+0.05} & \multirow{3}{*}{-} & \multirow{3}{*}{-} & \multirow{3}{*}{-} \\
& \multicolumn{1}{c|}{} & 13.2 \textcolor[rgb]{0.17,0.62,0.17}{(-0.2)} & 6.8 \textcolor{red}{(+0.6)} & & 26.5 \textcolor{red}{(+1.7)} & 24.4 \textcolor{red}{(+3.4)} & & 9.7 \textcolor[rgb]{0.17,0.62,0.17}{(-0.7)} & 10.8 \textcolor{red}{(+0.2)} & \textcolor{red}{+0.83} & & & \\
& \multicolumn{1}{c|}{} & \textbf{26.5 \textcolor{red}{(+1.4)}} & \textbf{29.2 \textcolor{red}{(+1.1)}} & & \textbf{43.5 \textcolor{red}{(+1.8)}} & \textbf{44.5 \textcolor{red}{(+1.5)}} & & 47.7 \textcolor{red}{(+0.9)} & \textbf{46.4 \textcolor{red}{(+1.0)}} & \textbf{\textcolor{red}{+1.28}} & & & \\ \hdashline[3pt/2pt]
\multirow{3}{*}{+ Seesaw Loss \cite{wang2021seesaw}} & \multirow{3}{*}{19.3} & 6.3 \textcolor{red}{(+1.5)} & 4.1 \textcolor{red}{(+0.6)} & \multirow{3}{*}{34.5} & 16.8 \textcolor{red}{(+2.6)} & 15.2 \textcolor{red}{(+1.4)} & \multirow{3}{*}{33.1} & 9.3 \textcolor{red}{(+0.5)} & 13.4 \textcolor[rgb]{0.17,0.62,0.17}{(-0.4)} & \textcolor{red}{+1.03} & \multirow{3}{*}{-} & \multirow{3}{*}{-} & \multirow{3}{*}{-} \\
& \multicolumn{1}{c|}{} & 12.8 \textcolor[rgb]{0.17,0.62,0.17}{(-0.6)} & 7.2 \textcolor{red}{(+1.0)} & & 26.7 \textcolor{red}{(+1.9)} & 24.7 \textcolor{red}{(+3.7)} & & 12.0 \textcolor{red}{(+1.6)} & 11.9 \textcolor{red}{(+1.3)} & \textcolor{red}{+1.48} &  & & \\
& \multicolumn{1}{c|}{} & 25.5 \textcolor{red}{(+0.4)} & 28.2 \textcolor{red}{(+0.1)} & & 43.0 \textcolor{red}{(+1.3)} & 44.2 \textcolor{red}{(+1.2)} & & 47.4 \textcolor{red}{(+0.6)} & 46.1 \textcolor{red}{(+0.7)} & \textcolor{red}{+0.71} & & & \\ \hdashline[3pt/2pt]
\multirow{3}{*}{+ FM (ours)} & \multirow{3}{*}{\textbf{20.3}} & \textbf{8.3 \textcolor{red}{(+3.5)}} & \textbf{7.6 \textcolor{red}{(+4.1)}} & \multirow{3}{*}{\textbf{36.3}} & \textbf{22.5 \textcolor{red}{(+8.3)}} & \textbf{19.5 \textcolor{red}{(+5.7)}} & \multirow{3}{*}{\textbf{34.0}} & \textbf{11.6 \textcolor{red}{(+2.8)}} & \textbf{17.0 \textcolor{red}{(+3.2)}} & \textbf{\textcolor{red}{+4.60}} & \multirow{3}{*}{-} & \multirow{3}{*}{-} & \multirow{3}{*}{-} \\
& \multicolumn{1}{c|}{} & \textbf{16.1 \textcolor{red}{(+2.7)}} & \textbf{8.3 \textcolor{red}{(+2.1)}} & & \textbf{31.7 \textcolor{red}{(+6.9)}} & \textbf{30.9 \textcolor{red}{(+9.9)}} & & \textbf{13.0 \textcolor{red}{(+2.6)}} & \textbf{12.3 \textcolor{red}{(+1.7)}} & \textbf{\textcolor{red}{+4.31}} &  & & \\
& \multicolumn{1}{c|}{} & 25.4 \textcolor{red}{(+0.3)} & 28.3 \textcolor{red}{(+0.2)} & & 42.4 \textcolor{red}{(+0.7)} & 43.6 \textcolor{red}{(+0.6)} & & \textbf{47.8 \textcolor{red}{(+1.0)}} & \textbf{46.4 \textcolor{red}{(+1.0)}} & \textcolor{red}{+0.63} & & & \\ \hline
\end{tabular}
}
\caption{\textbf{Benchmark of LTSS Datasets.} We evaluate out-of-the-box DeepLab-V3-Plus and mask2former with ResNet-50 backbone and several long-tailed solutions on three introduced LTSS datasets. Overall and image/pixel-level mIoU are reported to demonstrate performance on each frequency category: rare, common, and frequent.
 $\Delta_{freq}$ indicates the effects on rare, common, and frequent categories. Our proposed Frequency-based matcher (FM) achieves promising results, especially on rare and common categories. Image.: Image-level; Pixel.: Pixel-level. The top performance is highlighted in \textbf{bold} font.}
\label{table:baseline}
\end{table*}

\section{Experiments}
\label{sec:exp}
To help researchers understand the LTSS task more clearly and prove the effectiveness of the proposed FM, we conduct comprehensive experiments and analyses in this section.

\subsection{Implementation Details}
\noindent\textbf{Experimental Settings.} We leverage the out-of-the-box DeepLab-V3-Plus (OS16)\footnote{\fontsize{7pt}{1em}\url{https://github.com/facebookresearch/detectron2/tree/main/projects/DeepLab}} and mask2former\footnote{\fontsize{7pt}{1em}\url{https://github.com/facebookresearch/Mask2Former}} with the ResNet-50 \cite{he2016deep} backbone and conduct experiments on three constructed LTSS datasets (ADE20K-Full, COCO-Stuff-LT, MHP-v2-LT) and their balanced versions (ADE20K, COCO-Stuff, MHP-v2). We follow the hyperparameters of the original implementation. Each model is trained with 120 epochs. For data augmentation, we used random scales and horizontal flipping, followed by a fixed size crop to 512$\times$512 for all the experiments except mask2former with a Swin-L \cite{liu2021swin} of 640$\times$640. Test time augmentation techniques were not used. Unless otherwise stated, all the experiments used the same settings. The experiments were implemented in PyTorch and trained on 4 NVIDIA A100 GPUs with 40 GB of memory per card. To guarantee reproducibility, our code and models will be publicly available.

\noindent\textbf{Classical Solutions.}
We evaluate three classical long-tailed solutions (RFS, copy-paste, and seesaw loss) combined with mask2former on LTSS datasets. For fair comparisons, we search for the best hyperparameters for RFS and seesaw loss, and we release the code along with the specific configurations.

\noindent\textbf{Frequency-based Matcher.}
We studied the FM on three LTSS datasets with the mask2former and ResNet50 backbones. To obtain a unified form for all the datasets, $t$ is determined by the frequency threshold of the different frequency types. Without a specific statement, $t$ equals the maximum frequency of common classes in each dataset, and $s=2.0$.

\subsection{Main Results}
\label{sec:exp-ss}
\noindent\textbf{Semantic Segmentation Out-of-the-Box.} Table~\ref{table:baseline} shows the results of vanilla DeepLab-V3-Plus and mask2former with the ResNet-50 backbone on the LTSS and balanced datasets. It can be clearly observed that the performance on the LTSS datasets shows serious degradation compared with that on the balanced datasets, partly due to using less training data or more classes, but the essential reason is that the performance on these rare classes is seriously insufficient. Taking vanilla mask2former as an example, for the three LTSS datasets, the image-level mIoUs of rare classes are only 19.1\% (ADE20K-Full), 34.0\% (COCO-Stuff-LT) and 18.8\% (MHP-v2-LT) of the frequent classes. Moreover, this proportion also shows a positive correlation with the image-level Gini coefficient $\delta^\textit{image}$, \emph{i.e.}, the larger $\delta^\textit{image}$ is, the greater the accuracy gap between the accuracy of the frequent classes and the rare classes. Similar results can also be observed for the DeepLab-V3-Plus models.

\noindent\textbf{Classical Solutions on LTSS Datasets.} 
As shown in Table~\ref{table:baseline}, the three classical long-tailed solutions have some effect on the LTSS datasets but are not significant. Specifically, the improvement in RFS is not obvious except for the COCO-Stuff-LT dataset; however, it can improve the image-level mIoU of the rare classes by approximately 0.9\app1.9 points. Copy-paste has an increase of \app1 points mIoU on the three LTSS datasets.

However, this improvement does not seem to target rare classes but improves model generalizability through complex data augmentation. As a general training technique, copy-paste achieves promising overall mIoU values via mIoU$_f$ improvement. While the RFS and seesaw loss are designed for long-tailed problems, their effects are also unsatisfactory, since they inadvertently affect higher frequency categories during sampling process, such as frequent classes appearing alongside rare ones. Seesaw Loss, although it increases the weight of low-frequency classes, only affects the classification loss. Seesaw loss yields approximately 0.6\app1.1 point mIoU improvements, and the image-level mIoU$_\textrm{r}$ is improved by 0.5\app2.1 points. Similarly, for RFS, the average mIoU$_r$ improved by 1.08 points, and the average mIoU$_c$ improved by 0.76 points. The above results show that the classical long-tailed solutions have no obvious effect on the LTSS task. In particular, the improvement in the rare classes is insignificant.

\begin{table}[t]
	\centering
	\begin{threeparttable}
	\resizebox{0.5\textwidth}{!}{
	\setlength\tabcolsep{2.5pt}
        \renewcommand\arraystretch{1.2}
	\begin{tabular}{l|c|c|c}
	\hline\thickhline
	\rowcolor{mygray1}
        Methods & Long-tailed Mechanism & Backbone & mIoU  \\
	\hline
	\hline
	MaskFormer \cite{cheng2021perpixel}    & -                    & ResNet-50 & 17.4   \\
	Segmenter \cite{strudel2021segmenter}     & -               & ViT-B          & 17.9   \\
	RankSeg \cite{he2022rankseg}                 & -                  & ViT-B          & 18.7   \\ \hline
	\multirow{6}{*}{Mask2Former \cite{cheng2022masked}}              & -        & ResNet-50 & 18.8   \\
	 & RFS \cite{gupta2019lvis}   & ResNet-50  & 19.1   \\
 &  Copy-Paste \cite{ghiasi2021simple}   & ResNet-50 & 19.7   \\
 & Seesaw \cite{wang2021seesaw}   & ResNet-50 & 19.3   \\
  & X-Paste \cite{zhao2023x}   & ResNet-50 & 20.0   \\
  & Frequency-based Matcher (ours)       & ResNet-50  &  20.3   \\ \hdashline[3pt/2pt]
\multirow{4}{*}{Mask2Former \cite{cheng2022masked}} & -   & Swin-L & 25.6  \\
 & Copy-Paste \cite{ghiasi2021simple}   & Swin-L & 26.8   \\
& X-Paste \cite{zhao2023x}   & Swin-L & 27.0   \\
	      & Frequency-based Matcher (ours)       & Swin-L  & 27.4 \\ 
       \hdashline[3pt/2pt]
\multirow{3}{*}{Mask DINO \cite{li2023mask}} & -   & Swin-L & 26.0  \\
 & Copy-Paste \cite{ghiasi2021simple}   & Swin-L & 26.9   \\
  &  X-Paste \cite{zhao2023x}   & Swin-L & 27.3   \\
	      & Frequency-based Matcher (ours)       & Swin-L  &  \textbf{27.8} \\ 
	\hline
	\end{tabular}
	}
	\end{threeparttable}
	\caption{\textbf{Semantic segmentation performance comparison on ADE20K-Full}. The top performance is highlighted in \textbf{bold}. Our proposed FM method attains the best result of 27.8 mIoU. }
	\label{table:ade20k-full-comparison}
\vspace{-5mm}
\end{table}

\noindent\textbf{Frequency-based Matcher on LTSS Datasets.} At the bottom of Table~\ref{table:baseline}, we present the results of the FM. Our method significantly outperforms the baseline by 1.5-point mIoU on ADE20K-Full, 3.4-point mIoU on COCO-Stuff-LT, and 1.7-point mIoU on MHP-v2-LT. Notably, there is a remarkably greater boost in rare and common classes than in frequency classes. For example, on the COCO-Stuff-LT dataset, the image-level mIoU$_\textrm{r}$ is improved by 8.3 points, the mIoU$_\textrm{c}$ is improved by 6.9 points, and the mIoU$_\textrm{f}$ is improved by only 0.7 points. We also demonstrate the average deviation of the mIoU$_r$, mIoU$_c$, and mIoU$_f$ by $\Delta_{freq}$. We considerably increase average mIoU$_r$ by 4.6 points and average mIoU$_c$ by 4.31 points, while average mIoU$_f$ increases by 0.63 points.

Compared to classical solutions, our method not only achieves better overall performance but also significantly improves performance on low-frequency categories, which can be seen from $\Delta_{freq}$. We outperform the other methods by at least 3.52 points average mIoU$_r$ and 2.83 points $mIoU_c$, while the average mIoU$_f$ slightly decreases.

\noindent\textbf{Comparison to SOTA on ADE20K-Full.} As the ADE20K-Full dataset has been studied before, we compare its semantic segmentation performance with that of previous studies. The results are summarized in Table~\ref{table:ade20k-full-comparison}. We compare the FM algorithm with previous segmentation methods and our introduced long-tailed benchmark. To illustrate the potential and generality of this approach, we further conducted a stronger comparison with vanilla mask2former and mask2former + copy-paste employing the Swin-L backbone, as X-Paste achieved the best performance on ADE20K-Full among the classical solutions. As shown, mask2former achieves promising performance compared with the previous segmentation methods and reaches an mIoU of 18.8 points. Our frequency-based matcher achieves a 20.3\% mIoU with the ResNet-50 backbone, even exceeding the segmenter \cite{strudel2021segmenter} and RankSeg \cite{he2022rankseg}, which adopt a stronger backbone (ViT-B). Employing Swin-L and Mask DINO, our method attains the best result of 27.8\% mIoU and outperforms vanilla Mask DINO and Mask DINO + X-Paste by 1.8\% and 0.5\%, respectively.

\subsection{Ablation Study}
\begin{table}[t]
	\centering
	\begin{threeparttable}
	\resizebox{0.48\textwidth}{!}{
	\setlength\tabcolsep{2.5pt}
	\renewcommand\arraystretch{1.2}
	\begin{tabular}{c|c|c|ccc|ccc}
	\hline\thickhline
	\rowcolor{mygray1}
	                                       &                                           &                                                     &         \multicolumn{3}{c|}{Image-level}       &         \multicolumn{3}{c}{Pixel-level}         \\
	\cline{4-6} \cline{7-9}
	\rowcolor{mygray1}
        \multirow{-2}{*}{$t$}  & \multirow{-2}{*}{$s$} & \multirow{-2}{*}{mIoU} & mIoU$_\textrm{r}$ & mIoU$_\textrm{c}$ & mIoU$_\textrm{f}$ & mIoU$_\textrm{r}$ & mIoU$_\textrm{c}$ & mIoU$_\textrm{f}$  \\
	\hline
	\hline
	 \multicolumn{2}{c|}{baseline}                  & 18.8 & 4.8 & 13.4 & 25.1 & 3.5 & 6.2 & 28.1 \\
	\hline
	\multirow{4}{*}{0.0003} &       1.0                 & 18.6 & 6.0 & 11.6 & 24.6 & 3.6 & 7.7 & 27.1 \\
	\multirow{4}{*}{(rare)} &        2.0                  & 19.5 & 6.1 & 14.5 & 25.2 & 6.1 & 8.3 & 27.6 \\
	                                    &        \textbf{4.0}                & \textbf{20.1} & 7.9 & 14.8 & 25.7 & 7.5 & 8.5 & 28.0 \\
&        5.0                 & 20.0 & 7.8 & 14.2 & 26.0 & 7.1 & 8.3 & 28.7 \\
&        6.0                 & 19.8 & 7.2 & 14.0 & 25.6 & 6.8 & 7.9 & 28.0 \\

	\hline
	\multirow{2}{*}{0.0006} &       1.0                 & 19.2 & 6.8 & 16.1 & 24.1 & 4.8 & 8.7 & 27.4 \\
	\multirow{2}{*}{(common)}  &        \textbf{2.0}          & \textbf{20.3} & 8.3 & 16.1 & 25.4 & 7.6 & 8.3 & 28.3 \\
	                                    &        4.0                 & 20.1 & 7.0 & 15.2 & 25.9 & 5.3 & 7.8 & 29.1 \\
	\hline
	\end{tabular}
	}
	\end{threeparttable}
	\caption{\textbf{Influence of $t$ and $s$ of the frequency-based matcher on ADE20K-Full}. 
 ``baseline'' refers to the vanilla mask2former.}
	\label{table:freq-based-matcher-ablation}
\end{table}

\begin{figure*}[!h]
	\begin{center}
		\includegraphics[width=0.99\linewidth]{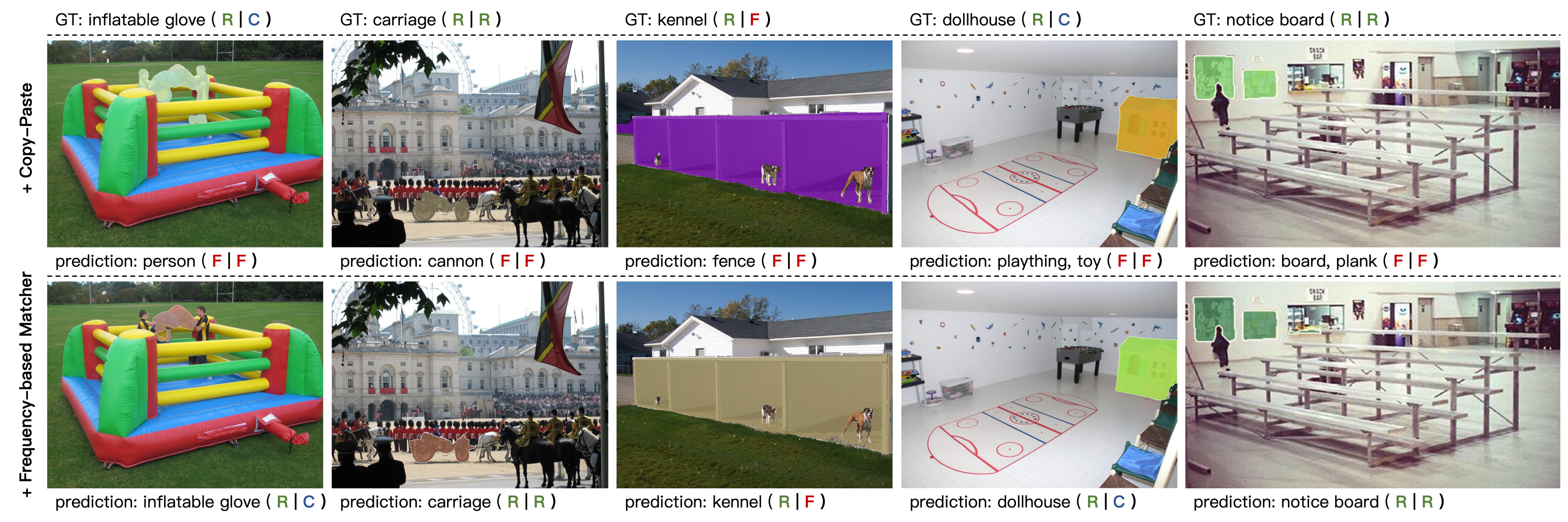}
	\end{center}
	\vspace{-4mm}	
	\caption{\textbf{Visualization of rare category predictions.} We further show the image-level and pixel-level frequencies of each category. The frequent, common, and rare categories are labeled \textcolor[rgb]{0.75,0,0}{F}, \textcolor[rgb]{0.18,0.33,0.59}{C}, and \textcolor[rgb]{0.32,0.50,0.20}R, respectively.}
	\label{fig:comparison}
\end{figure*}

We studied the hyperparameters of ADE20K-Full with the mask2former and ResNet50 backbones. As previously mentioned, $t$ acts as a frequency threshold to control the ratio of categories matching more than one query, and $s$ controls the degree of extra queries. A higher $t$ increases the number of classes applied to one-to-many matching. A higher $s$ increases the number of matched queries for all the classes. The results are shown in Table~\ref{table:freq-based-matcher-ablation}.

For ADE20K-Full, 0.0003 and 0.0006 are frequency quartiles of rare and common categories, respectively. For example, $t=0.0003$ means that all rare classes match more queries. We find that $t=0.0006$ and $s=2.0$ can achieve the best performance, with an overall mIoU of 20.3 points, which is 1.5 points greater than that of the vanilla mask2former. Specifically, an 8.3-point mIoU$_r$ is achieved at the image level, and a 7.6-point mIoU$_r$ is achieved at the pixel level. A FM significantly benefits rare classes with various settings and increases 1.2$\sim$3.5 points mIoU$_r$ at the image level and 0.1$\sim$4.1 points mIoU$_r$ at the pixel level since rare classes can match more queries than the baseline.

When $t=0.0003$, only rare classes have a relative matching intensity. Therefore, the mIoU$_\textrm{r}$ increases significantly, and the mIoU$_\textrm{c}$ increases slightly. When $t$ increases to 0.0006, common classes also match more queries, and the mIoU$_\textrm{c}$ further improves. As $s$ increases, the overall mIoU improves and saturates when $s$ increases. Regardless of how $t$ and $s$ change, the impact on the mIoU$_\textrm{f}$ is small, which also shows that our FM enhances the low-frequency class.


\subsection{Analysis and Discussion}
In this subsection, we conduct an overall analysis and discussion of the FM.

\noindent\textbf{Qualitative Results.} We further visualize some results of copy-paste and the FM under mask2former and Swin-L in Figure~\ref{fig:comparison}. Copy-paste prefers to misclassify rare classes as frequent classes. Benefiting from frequency-based supervision enhancement, our method predicts rare classes more accurately.

\begin{table}[t]
	\centering
	\begin{threeparttable}
	\resizebox{0.5\textwidth}{!}{
	\setlength\tabcolsep{2.5pt}
        \renewcommand\arraystretch{1.2}
	\begin{tabular}{l|c|c|cc}
	\hline\thickhline
	\rowcolor{mygray1}
        Methods & \tabincell{c}{Frequency-based \\Matcher} & mIoU  & mIoU$^\textit{image}_\textrm{r}$  & mIoU$^\textit{pixel}_\textrm{r}$ \\
	\hline
	\hline
        \multirow{2}{*}{Mask2Former \cite{cheng2022masked}}    & - & 18.8 & 4.8 & 3.5 \\
                                                               & \checkmark & 20.3  & 8.3 & 7.6   \\
        \hdashline[3pt/2pt]
	\multirow{2}{*}{+ RFS \cite{gupta2019lvis}}            & -          & 19.1  & 5.7 & 4.1  \\
                                                               & \checkmark & 20.7  & 9.5 & 8.0   \\
        \hline
	\multirow{2}{*}{+ Copy-Paste \cite{ghiasi2021simple}}  & -          & 19.7  & 4.8 & 3.7   \\
                                                               & \checkmark & 21.2  & 8.5 & 7.4    \\
        \hline
	\multirow{2}{*}{+ Seesaw Loss \cite{wang2021seesaw}}    & -         & 19.3  & 6.3 & 4.1   \\
                                                               & \checkmark & 20.4  & 10.0 & 8.1   \\
	\hline
 \multirow{2}{*}{+ X-Paste \cite{zhao2023x}}    & -         & 20.0  & 6.7 & 4.4   \\
                                                               & \checkmark & 21.6  & 9.8 & 8.3   \\
	\hline
	\end{tabular}
	}
	\end{threeparttable}
	\caption{\textbf{The frequency-based matcher collaborates with different long-tailed learning strategies on ADE20K-Full}. Notably, mIoU$^\textit{image}_\textrm{r}$ denotes the mIoU$_\textrm{r}$ at image-level, and mIoU$^\textit{pixel}_\textrm{r}$ denotes the mIoU$_\textrm{r}$ at pixel-level.}
	\label{table:collaborating-others}
\vspace{-5mm}
\end{table}

\noindent\textbf{Collaborating with other strategies.} Since other long-tailed learning strategies do not focus on the matcher, we consider that FMs can collaborate with them to achieve greater performance. As shown in Table~\ref{table:collaborating-others}, our method can further improve segmentation performance on other long-tailed learning strategies. Among them, the combination of the FM and X-Paste achieved the highest overall mIoU, with an improvement of 2.8 points compared to the vanilla mask2former baseline and 1.6 points compared to the X-Paste-only strategy. In addition, the combination of FM and seesaw loss has the best performance in low-frequency category segmentation, which outperforms the vanilla mask2former baseline with 5.2-point mIoU$^\textit{image}_\textrm{r}$ and 4.6-point mIoU$^\textit{pixel}_\textrm{r}$.

\noindent\textbf{Integration with Other Transformer-based Segmenters.} The FM improves the matching strategy between queries and ground-truth masks, allowing for more sufficient supervision of low-frequency class instances and making it suitable for the vast majority of transformer-based segmenters, \emph{e.g.}, Group DETR \cite{chen2022group}, Mask DINO \cite{li2023mask}, and MP-Former \cite{zhang2023mpformer}. Table~\ref{table:other-segmenters} shows the performances of three transformer-based segmenters with and without the FM on ADE20K-Full. It can be clearly observed that the FM improves the performance of each segmenter (+1.2 \app +1.8 mIoU). Specifically, for rare classes, the effect of the FM is more significant for both the image-level and pixel-level metrics, indicating that the FM is a universal LTSS strategy that can seamlessly improve the performance of transformer-based segmenters for rare objects or stuff.

\begin{table}[t]
	\centering
	\begin{threeparttable}
	\resizebox{0.5\textwidth}{!}{
	\setlength\tabcolsep{2.5pt}
        \renewcommand\arraystretch{1.2}
	\begin{tabular}{l|c|c|cc}
	\hline\thickhline
	\rowcolor{mygray1}
        Methods & \tabincell{c}{Frequency-based \\Matcher} & mIoU  & mIoU$^\textit{image}_\textrm{r}$  & mIoU$^\textit{pixel}_\textrm{r}$ \\
	\hline
	\hline
	\multirow{2}{*}{Group DETR \cite{chen2022group}}       & -                & 19.2  & 4.9 & 3.7  \\
                                                               & \checkmark      & 21.0  & 9.2 & 8.7   \\
        \hline
	\multirow{2}{*}{Mask DINO \cite{li2023mask}}           & -                 & 20.1  & 5.7 & 4.8   \\
                                                               & \checkmark       & 21.3  & 9.4 & 9.6    \\
        \hline
	\multirow{2}{*}{MP-Former \cite{zhang2023mpformer}}    & -                  & 18.9  & 5.1 & 3.3   \\
                                                               & \checkmark       & 20.2  & 8.7 & 6.9   \\
	\hline
	\end{tabular}
	}
	\end{threeparttable}
	\caption{\textbf{Performance of three transformer-based segmenters on ADE20K-Full with/without the FM}. We adopt the default configuration from the original paper for each segmenter.}
	\label{table:other-segmenters}
\vspace{-5mm}
\end{table}

\section{Conclusion and future work} 
\label{sec:conc}
In this work, we focus on a challenging problem, \emph{i.e.}, long-tailed semantic segmentation (LTSS), which aims to learn dense pixel prediction that can account for both head and tail classes from a long-tailed distribution. We established a complete LTSS learning system around this novel task setting, including three datasets with multiple data scenarios, a dual-metric evaluation system, and a solid benchmark based on an advanced semantic segmentation pipeline. After deeply analyzing the bottleneck of the classical long-tailed learning solutions in the LTSS task, we propose a transformer-based targeted approach, the FM, which solves the oversuppression problem by one-to-many matching and automatically determines the number of matching queries for each class. The experimental results on the ADE20K-Full, COCO-Stuff-LT, and MHP-v2-LT datasets demonstrate the superiority of our proposed approach. Our proposed FM represents a significant advance in transformer-based frameworks for long-tailed semantic segmentation, yet there are key areas where further development is needed. A primary limitation is that FM is only compatible with transformer-based frameworks, although they lead in segmentation accuracy. Additionally, there is a rich avenue for research in other LTSS aspects, such as data augmentation, which can offer substantial performance benefits. Future work will delve into these areas, seeking comprehensive and effective solutions that enhance the robustness and versatility of LTSS approaches. 

\section{Acknowledgments} 
This work was supported by China National Postdoctoral Program for Innovative Talents (Grant No. BX2021047), China Postdoctoral Science Foundation (Grant No. 2022M710466), Young Scientists Fund of NSFC (Grant No. 62206025), and Program for Youth Innovative Research Team of BUPT (No. 2023QNTD02).

	{
\bibliographystyle{IEEEtran}
\bibliography{IEEEabrv,egbib}
	}
	
\end{document}